# HE-SLAM: a Stereo SLAM System Based on Histogram Equalization and ORB Features

Yinghong Fang [a], Guangcun Shan [*,a], Xin Li [a], Wenliang Liu [a], Tian Wang [b], Hichem Snoussi [c],

[a] *School of Instrumentation Science and Opto-electronic Engineering, Beihang University*, Beijing, China
[b] *School of Automation Science and Electrical Engineering, Beihang University*, Beijing, China
[c] *Institute Charles Delaunay-LM2S-UMR STMR 6279 CNRS, University of Technology of Troyes,* Troyes, France

*Abstract*—In the real-life environments, due to the sudden appearance of windows, lights, and objects blocking the light source, the visual SLAM system can easily capture the low-contrast images caused by over-exposure or over-darkness. At this time, the direct method of estimating camera motion based on pixel luminance information is infeasible, and it is often difficult to find enough valid feature points without image processing. This paper proposed HE-SLAM, a new method combining histogram equalization and ORB feature extraction, which can be robust in more scenes, especially in stages with low-contrast images. Because HE-SLAM uses histogram equalization to improve the contrast of images, it can extract enough valid feature points in low-contrast images for subsequent feature matching, keyframe selection, bundle adjustment, and loop closure detection. The proposed HE-SLAM has been tested on the popular datasets (such as KITTI and EuRoc), and the real-time performance and robustness of the system are demonstrated by comparing system runtime and the mean square root error (RMSE) of absolute trajectory error (ATE) with state-of-the-art methods like ORB-SLAM2.

## I. Introduction

Recently, Simultaneous Localization and Mapping (SLAM) has developed into a research hotspot in the field of mobile robots and is considered to be the core link to achieve autonomous navigation [1]. SLAM is defined as: in the unknown environment, the robot uses sensor information to realize pose estimation and incremental construction of environment map and uses pose information and map model to achieve global positioning.

At present, the conventional robot SLAM systems generally have two forms: LiDAR-based SLAM (laser SLAM) and vision-based SLAM (Visual SLAM or VSLAM) [2]. In this paper, we concentrate on the latter, which consists of sensor information input, visual odometry (VO, also known as front end), optimization (because after the VO, also known as back end), loop closing and mapping [3]. The goal of the visual odometer is to estimate the motion of the camera based on the captured image. Its main ways are divided into the feature point method and the direct method. The direct methods (typical algorithms such as DTAM [4], LSD-SLAM [5], and DSO [6], etc.) estimate the camera motion according to the pixel luminance information of the image and optimizes it by minimizing the luminosity error. However, it is based on the gray-scale invariant assumption, that is, the pixel gray level of the same spatial point is fixed in each image. For the sudden appearance of windows, lights and objects blocking the light sources described in this paper, the conditions of utilizing the direct method cannot be satisfied, and therefore it is necessary to adopt the feature point method.

However, if the features are directly extracted from the low-contrast images with over-exposure or over-darkness, it is difficult to obtain enough valid feature points. Therefore, HE-SLAM is proposed in this work, for low-contrast images, the histogram equalization is first performed to improve the contrast of images, and then the feature points are extracted. To enhance the real-time performance of feature extraction and matching, ORB (oriented FAST and rotated BRIEF) [7] feature extraction and matching algorithm is studied and improved. After that, keyframe selection, bundle adjustment and loop closure detection are performed. Finally, the SLAM framework is tested on standard datasets to verify that it can establish an environmental model with global consistency and global estimation, and the real-time performance and robustness of the improved algorithm and system framework are also evaluated in this paper.

## II. HE-SLAM Overview

The method proposed in this work contains several modules in the classic visual SLAM: the front end (visual odometry) incrementally solves the pose by matching the point features; the back end receives the posture of the visual odometry estimated at different moments, and the information of loop closure detection, and optimizes them to obtain a globally consistent trajectory and map.

---

[*] Address all correspondence to this author. Email: gcshan@buaa.edu.cn;

A general process of the HE-SLAM system is displayed in Fig. 1. The system is extended by ORB-SLAM [8] and adopts the same primary structure. The system is mainly divided into three different threads: visual odometry, local mapping, and loop closing. This efficient distribution allows for continuous tracking of VO modules, while local mapping and loop closing are only processed in the background when new keyframes are inserted.

The main contents of each thread are given as follows:

1) Visual odometry: input the image sequence of stereo frames, which is divided into left images and right images. The left and right images at the same time are called one frame. Image preprocessing includes histogram equalization, distortion correction, detection and description of feature points, and stereo matching. Tracking is divided into two stages, one is to track the adjacent frames, the other is to follow the local map, and the pose of the camera is obtained by minimizing the re-projection error. Finally, the key frames of the current frame are judged [9].

2) Local mapping: optimizes the points and poses in the local map after the tracking thread inserts keyframes [10]. At the same time, according to the statistical information, the spatial points in the map are eliminated, the stable tracking part is retained, and the keyframes with redundant information in the map are eliminated. After insertion of keyframes, new map points will be created combined with another frame in the local map.

3) Loop closing: loop closure detection is performed by a bag of words (BoW) [11] approach. When the closed loop is detected, the SE (3) transformation between the closed-loop frame and the current frame is calculated, and the accumulative errors and the position of map points are corrected by optimizing the pose map [12].

In the present work, we focus on visual odometry.

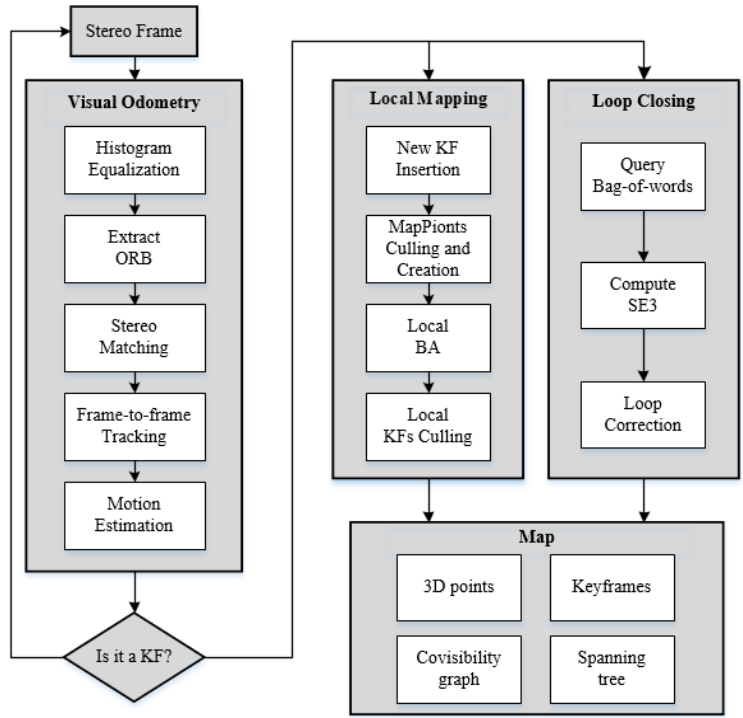

Fig. 1. Scheme of the HE-SLAM system.

### III. HISTOGRAM EQUALIZATION

At present, the commonly used method of image contrast enhancement is histogram equalization [13]. Its basic idea is to transform the histogram of the original image into a uniform distribution form, thus increasing the dynamic range of pixel gray pixel, thereby enhancing the overall contrast of the image, which is beneficial for the ORB algorithm to extract image features. The following is the principle of histogram equalization:

Let the variable $r$ represent the gray level of the pixels in the image. Normalizing the gray level, $0 \leq r \leq 1$, where $r = 0$ for black and $r = 1$ for white. For a given image, each pixel value is random at the gray level of [0,1]. The probability density function $p_r(r)$ is used to represent the distribution of gray levels.

To make digital image processing easier, we introduce discrete forms. In discrete form, $r_k$ is used to represent the discrete gray level, and $p_r(r_k)$ is used to represent $p_r(r)$, and the following formula holds:

$$p_r(r_k) = \frac{n_k}{n} \tag{1}$$

where $0 \leq r_k \leq 1$, $k = 0, 1, 2, ..., n$-1. In the formula, $n_k$ is the number of pixels in the image where $r_k$ appears, $n$ is the total number of pixels in the image, and $n_k/n$ is the frequency in probability theory. The function expression for the histogram equalization of the image is:

$$s_k = T(r_k) = \sum_{i=0}^{k} p_r(r_i) = \sum_{i=0}^{k} \frac{n_i}{n} \tag{2}$$

where $k$ is the number of gray levels. Therefore, the gray level in the input image is mapped from each pixel value of $r_k$ to a corresponding pixel value with a gray level $s_k$.

In the specific algorithm implementation steps, for the grayscale image with the grayscale value between 0 and 255, a further action is needed to calculate the pixel value $y_k$ of the transformed image:

$$y_k = 255 \times s_k \tag{3}$$

## IV. ORB Feature Point Detection and Matching

The feature points of an image can be simply understood as points that are more prominent in the images, such as contour points, bright points in darker areas, dark spots in brighter areas, etc. In the SLAM system framework that deals with scene transformation, feature points should have the better scale, rotation invariance and real-time performance. The ORB feature consists of two parts: a key point and a descriptor. The ORB image feature matching algorithm uses oFAST corner detection [14] and rBRIEF feature descriptor to achieve image feature matching.

### A. FAST Algorithm and Improvement

ORB algorithm uses the FAST (features from accelerated segment test) algorithm to detect feature points. The FAST algorithm detects the pixel value of a circle around a candidate feature point based on the gray value of the image around the feature point. If enough pixels in the field around the candidate point differ from the gray value of the candidate point sufficiently, the candidate point is considered to be a feature point.

As the feature detection part of ORB algorithm, the FAST algorithm can improve the efficiency and quality of feature detection. The feature points obtained by FAST corner detection algorithm do not have direction information, so the oFAST (oriented FAST) descriptor with rotation invariance can be obtained by specifying intensity centroid as the rotation direction.

The intensity centroid method assumes that the gray centroid of the feature points deviates from the center of gravity of the region in which they are located. According to the deviation, the vector can be defined as the direction of the feature points. The process is to calculate the centroid position $C$ in the neighborhood $S$, take the feature point $O$ as the origin of the coordinate system, and construct the vector $\overrightarrow{OC}$ with the center of gravity as the endpoint.

The moment of neighborhood S can be expressed as:

$$m_{pq} = \sum_{x,y} x^p y^q I(x,y) \tag{4}$$

Where $I(x,y)$ represents the gray value of the image point $(x,y)$, $x, y \in [-r, r]$, and $r$ is the radius of the neighborhood $S$. Let $p$ and $q$ be 0 and 1, respectively, to obtain the grayscale centroid of the region I in the $x$, $y$ direction. Then the centroid position of this neighborhood is:

$$C = (\frac{m_{10}}{m_{00}}, \frac{m_{01}}{m_{00}}) \tag{5}$$

Finally, the direction of FAST feature points can be obtained:

$$\theta = \operatorname{atan2}(m_{01}, m_{10}) \tag{6}$$

### B. BRIEF Feature Descriptor Algorithm

BRIEF is a local feature descriptor in the form of binary encoding. The basic idea of BRIEF is to select several pairs of pixels $p$, $q$ in the neighborhood block of feature points according to specific rules. If $I_p > I_q$, the value of the BRIEF descriptor of the current bit is 1, otherwise it is 0. For comparing the number of pixel pairs, you can usually choose 128 or 256.

But in fact, there is a very serious problem, that is, the obtained descriptors will be very sensitive to noise. If the pixels are compared to noise, they will have a significant impact on the reliability of the descriptors. Therefore, the above comparison rules have been adjusted:

$$\tau(p;x,y) := 1 : p(x) > p(y); \tau(p;x,y) := 0 : p(x) \leq p(y); \quad (7)$$

Where *p* (*x*) represents the intensity value of pixel *x*, and the Gaussian filter value can be selected as the intensity value of the current pixel point, so that the two compared values are not two separate pixel point values, but the weighted average value of neighborhood blocks of pixel points, thus effectively reducing the impact of noise.

The BRIEF descriptor itself does not solve the problem of rotation invariance, so in the ORB features to solve this problem, the solution is also straightforward, that is, in the process of calculating feature point descriptors, correlating some comparisons of the neighborhoods with corresponding rotations according to the direction of the feature points, that is, if n test pairs' coordinates are represented as:

$$S = (P_1, P_2, \ldots, P_n), P_i = (x_i, y_i) \quad (8)$$

The direction angle of the feature point is *θ*, and the corresponding rotation matrix is $R_\theta$. Then the revised comparison pair coordinate $S_\theta$:

$$S_\theta = R_\theta S \quad (9)$$

The modified coordinate $S_\theta$ is brought into the calculation of the descriptor, so that the resulting descriptor is rotationally invariant, called rBRIEF.

The Hamming distance can be used to determine whether the rBRIEF feature points are matched. The generated rBRIEF feature point descriptor is a binary code string, and the Hamming distance is established by the XOR operation of the binary line. Therefore, the feature matching process is fast and straightforward. The fast library for approximate nearest neighbors (FLANN) algorithm can be used to match the rBRIEF feature description. It is generally considered that the smaller the Hamming distance between the current feature point and the nearest neighbor point, and the larger the Hamming distance between the current feature point and the next nearest neighbor point, the better the matching quality of the current rBRIEF is.

TABLE I. ATE RMSE IN THE KITTI DATASET

| Seq. | ATE RMSE (m) | |
|---|---|---|
| | *HE-SLAM* | *ORB-SLAM2* |
| 00 | **1.3034** | 1.30345 |
| 01 | **10.2217** | 10.8385 |
| 02 | 5.74836 | **5.6336** |
| 03 | 0.996917 | **0.977166** |
| 04 | 0.233628 | **0.163535** |
| 05 | 0.805461 | **0.764564** |
| 06 | **0.812416** | 1.05377 |
| 07 | 0.500916 | **0.489687** |
| 08 | **3.08941** | 3.40213 |
| 09 | **3.61144** | 3.87987 |
| 10 | **0.924477** | 1.02258 |

## V. THE BACK END OF SLAM SYSTEM

The back end of the system is responsible for the optimization of pose map to make the output map model globally consistent [15]. Factors such as sensor noise, cumulative error in depth value quantization and registration errors will cause the system to be uncertain, resulting in shifts in pose and 3D maps.

To enhance the global consistency of the generated map model, we use an optimized back-end, consisting of local bundle adjustment, loop closure detection and trajectory optimization, to update the whole map in real-time during the process of map construction.

## VI. EVALUATION

Our HE-SLAM system runs on the Ubuntu 14.04 platform configured with Intel Core i7-8750H CPU@2.20GHz and 8 GB RAM.

HE-SLAM has been tested on the popular datasets (such as KITTI and EuRoc), and the real-time performance and robustness of the system are demonstrated by comparing system runtime and the mean square root error (RMSE) of absolute trajectory error (ATE) with state-of-the-art methods like ORB-SLAM2.

## A. Kitti Dataset

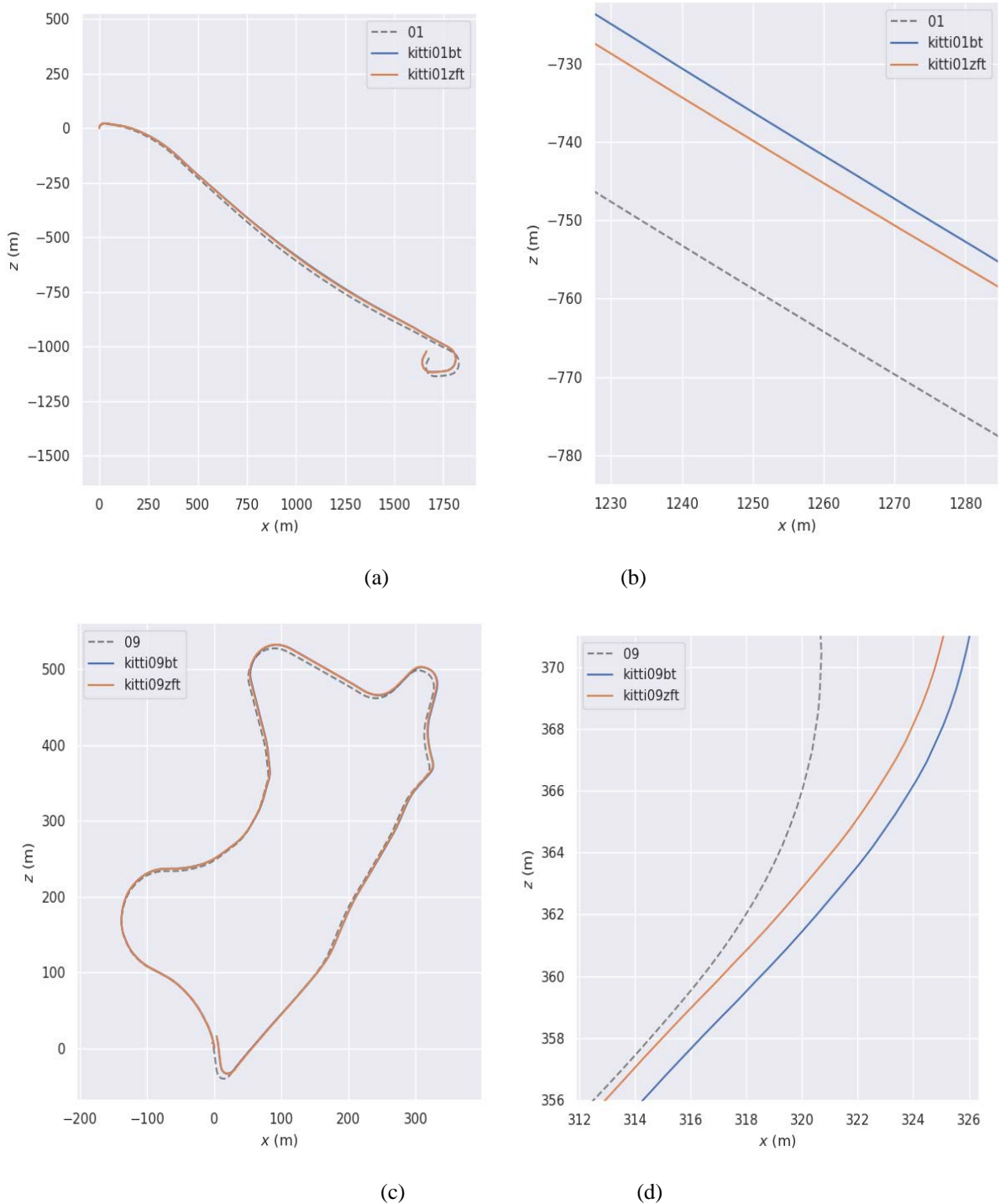

(a) (b)

(c) (d)

Fig. 2. Dotted lines represent ground truth, blue Line represents ORM-SLAM2 trajectory and red Line represents HE-SLAM trajectory. (a)(b) show the trajectory in the sequence KITTI-01, (b) is a locally magnified graph of (a). (c)(d) show the path in the sequence KITTI-09, (d) is a locally expanded graph of (c).

## B. EuRoc MAV Dataset

TABLE II. ATE RMSE IN THE EUROC MAV DATASET

| Sequence | ATE RMSE(m) | |
| --- | --- | --- |
| | *HE-SLAM* | *ORB-SLAM2* |
| MH-01-easy | 0.0410544 | **0.0380801** |
| MH-02-easy | **0.0404337** | 0.0436409 |
| MH-03-med | 0.0507298 | **0.0405288** |
| MH-04-dif | **0.105072** | 0.121237 |
| MH-05-dif | **0.0520933** | 0.0970062 |
| V1-01-easy | **0.0884811** | 0.0885614 |
| V1-02-med | 0.0643252 | **0.0630205** |
| V1-03-dif | 0.0809867 | **0.0650267** |
| V2-01-easy | 0.0625152 | **0.0584563** |
| V2-02-med | 0.0639234 | **0.0573975** |
| V2-03-dif | **0.234987** | 0.236151 |

It is found that, in table I, the sequence where the ATE RMSE of ORB-SLAM2 is very large, the ATE RMSE of HE-SLAM is Smaller, the map is simple, as shown in Fig.2.

This phenomenon means that ORB-SLAM2 is suitable for a larger scene and closed-loop environment, so the camera pose and map can be optimized by loop closure detection. But HE-SLAM has advantages in the absence of loop closure detection in small scenes, and it takes a little less time (Time have been measured, it is basically flat so it is not shown). And the phenomenon in Kitti dataset is also present in the EuRoc dataset. (In addition to the sequence V2-03-difficult, the image set of the sequence itself is problematic.) As a result, our HE-SLAM Slam is more robust than ORB-SLAM2 and can also be used in harsh environments.

## VII. CONCLUSION

Considering the sudden appearance of windows, lights and objects blocking light sources in real life, the visual slam system can easily capture images with low contrast caused by over-exposure and over-darkness. A novel stereo visual SLAM system has been proposed, coined HE-SLAM, which can be real-time and robust in many types of environments, including low-contrast ones. Our proposed HE-SLAM system uses histogram equalization to improve the contrast of the image, which can extract enough useful feature points in low-contrast pictures for subsequent feature matching, keyframe selection, bundle adjustment, and loop closure detection. The real-time performance and robustness of the HE-SLAM system have been evaluated by the RMSE and system running time. The experimental results based on the standard test datasets (such as KITTI and EuRoc) have shown that our HE-SLAM proposed here can effectively improve the speed and stability of the SLAM system.

## ACKNOWLEDGMENT

This work was supported by the National Key R&D Program of China (Grant No. 2016YFE0204200).